# An Algorithm for Generating Gap-Fill Multiple Choice Questions of an Expert System

Pornpat Sirithumgul
Department of Computer Engineering
Rajamangala University of Technology
Phra Nakhon
pornpat.s@rmutp.ac.th

Pimpaka Prasertsilp
School of Science and Technology
Sukhothai Thammathirat Open
University
pimpaka.pra@stou.ac.th

Lorne Olfman
Center for Information Systems and
Technology
Claremont Graduate University
lorne.olfman@cgu.edu

**Abstract**

*This research is aimed to propose an artificial intelligence algorithm comprising an ontology-based design, text mining, and natural language processing for automatically generating gap-fill multiple choice questions (MCQs). The simulation of this research demonstrated an application of the algorithm in generating gap-fill MCQs about software testing. The simulation results revealed that by using 103 online documents as inputs, the algorithm could automatically produce more than 16 thousand valid gap-fill MCQs covering a variety of topics in the software testing domain. Finally, in the discussion section of this paper we suggest how the proposed algorithm should be applied to produce gap-fill MCQs being collected in a question pool used by a knowledge expert system.*

## 1. Introduction

This research is motivated by the use of online learning, e.g., Massive Open Online Courses (MOOCs) [1], and Coursera [2], which has been increasing in academia. One key limitation is that online learning has not yet had an automatic question generator for assessing knowledge, especially in the scientific areas. This current research study is aimed to reduce the limitation by proposing an algorithm as a foundation for developing software that can automatically generate gap-fill multiple choice questions (MCQs) for measuring knowledge in any given scientific domain. In this research, we define a gap-fill multiple choice question (MCQ) as a fill-in-the-bank question where one keyword in the question phrase is deprived, and listed in the set of four potential answer choices.

There are several preliminary studies [3 - 13] proposing methods for automatically generating questions used in computer-based offline/online learning. These studies, however, have not reported success of applying the methods in the actual learning setting.

A number of early research studies in this area [14 - 18] focus on automatic generation of MCQs to assess knowledge in the linguistic domain. There are several attempts [19 - 37] at applying MCQ generation methods in the areas of generic sciences [19 - 22], biology and medical sciences [23 - 29], technology [30 - 36], and computer programming [37]. These studies, however, have not reported whether the generated MCQs thoroughly covered all topics in the knowledge domains.

To fulfill the gap found in this research area, we propose an algorithm for automatically producing gap-fill MCQs covering topics in a given scientific domain. This current research study also demonstrates a usage of the proposed algorithm in the domain of Software Testing. The simulation results from the study indicate that the algorithm can be generalized to produce gap-fill MCQs thoroughly covering topics in any given scientific domain.

## 2. Knowledge Background

The algorithm proposed in this research comprises three foundation theories: ontology-based design, text mining, and natural language processing. Details of the three base theories are as follows.

### 2.1. Ontology-based Design

**Ontology-based design** is a part of the algorithm focusing on specifying concepts and concepts' relations in the scope of a knowledge domain. By following the previous research [38 - 39], concepts' relations (see Figure 1) are defined in four classes including association, generalization, composition, and aggregation.

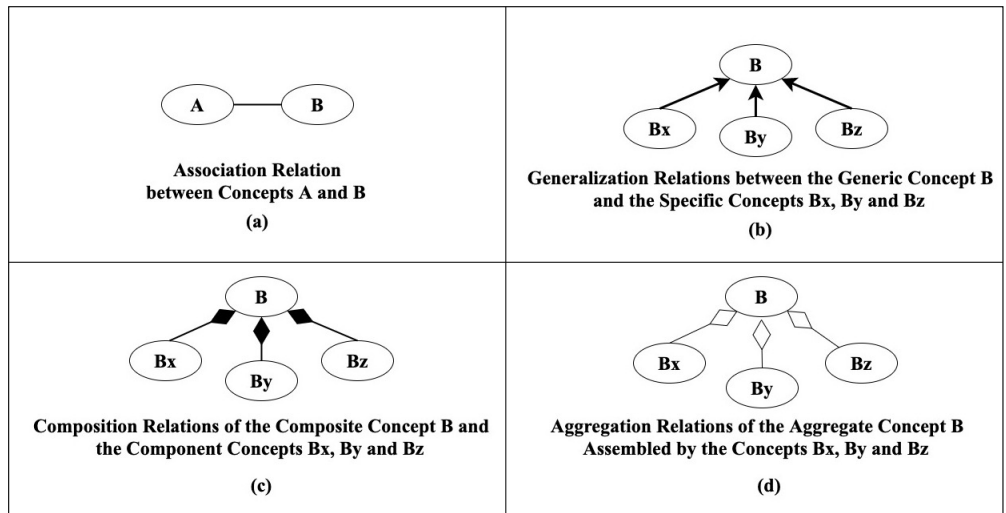

**Figure 1.** Four Classes of Concepts' Relations

Association as symbolized in Figure 1 (a) is a relation basically used to describe mutual concepts frequently presented together in the real world, e.g., paper and pens, or birthday cake and candles. Generalization as in Figure 1 (b) is used to describe a relation between a generic idea and specific ideas or instances of the generic idea. For example, *'hospital personnel'* is a generic idea about people working in a hospital, while the instances – medical doctors and nurses represent the personnel having particular tasks in the hospital. Composition is a relation describing components and a collective piece of the components, e.g., wheels and a car. Aggregation is similar to composition; it describes components and an aggregate concept. However, both component and aggregate concepts can stay independent, e.g., students aggregating into a class.

## 2.2. Text Mining

***Text mining*** especially for this research is the process of finding technical terms describing concepts and concepts' relations in the specified knowledge domain. Referring to the previous research [38 - 39], technical terms representing focal concepts in a domain are initially used to find other co-occurring technical terms. *'Unique term'*, *'frequent term'*, and *'common term'* are the three classes of the co-occurring terms first defined in [38]. A unique term means a technical term co-occurring with the term representing a concept, and being presented only in the documents specifically describing the concept. A frequent term means a technical term co-occurring with the term representing a concept, and being presented frequently in the documents specifically describing the concept, and occasionally in the documents describing other relative concepts. A common term means a technical term commonly co-occurring with some other technical terms in varied documents describing relative concepts in the same knowledge domain.

This current research uses text mining to filter common terms out of unique and frequent terms that take the role of keywords describing the essential characteristics of focal concepts. Further, the algorithm proposed in this research would use unique and frequent terms to spot sentences containing the terms, and later use the sentences to produce question phrases in gap-fill MCQs.

## 2.3. Natural Language Processing

Natural Language Processing (NLP) is used as a part of the algorithm for preparing data, and for examining sentence structures. Lemmatization, tokenization, and part-of-speech tagging are the specific techniques of NLP included in the algorithm. Details of the three techniques follow.

**2.3.1. *Lemmatization*** is the technique for converting words to their dictionary form [40 - 41]. A conversion, for example, is from a plural to singular noun, or from a present or past participle verb to a simple present verb, e.g., converting *'walking'* and *'walked'* to *'walk'*. Especially for this current research, lemmatization is used to change a technical term to its original form as appearing in the glossary, e.g., *'test suites'* would be changed to *'test suite'*.

**2.3.2. *Tokenization*** is the technique for breaking up a string of words [40 - 41], i.e., breaking a phrase or a sentence into a set of independent tokens. For example, a set of tokens of the sentence *"My father is traveling to London."* is {*'My'*, *'father'*, *'is'*, *'traveling'*, *'to'*, *'London'*, *'.'*}. However, tokenization is not done easily by using spaces in splitting words because a compound word, e.g., *'living room'* should be preserved rather than split into two tokens – *'living'* and *'room'* that convey a different meaning. Particularly for this current research, a glossary is used to look up compound technical terms, e.g., *'cyclomatic complexity'*, *'control flow graph'*, that would be preserved in the tokenizing process.

**2.3.3.** *Part-of-Speech (POS)* tagging is the technique for classifying and tagging words by their parts of speech [40 - 41]. The main parts of speech include noun, pronoun, verb, adjective, adverb, preposition, conjunction, and interjection. The POS tagging technique is incorporated into the algorithm of this current research in order to examine a structure of sentences suitable for using as question phrases in gap-fill MCQs, that is, the required sentences must not have an *adverb* at the beginning, and must be in the form of *subject + verb*, or *subject + verb + object*.

## 3. Proposed Algorithm

The algorithm proposed in this research comprises three main steps (see Figure 2). Step 1, called *'Corpus Management'*, is for building a corpus specifically used for collecting technical terms in a given knowledge domain. Step 2, called *'Sentence Selection'*, is for selecting sentences suitable for being used as question phrases. Lastly, Step 3, called *'Gap-Fill MCQ Generation'*, is for formulating complete gap-fill MCQs by using the technical terms from the corpus, and the sentences collected at Step 2.

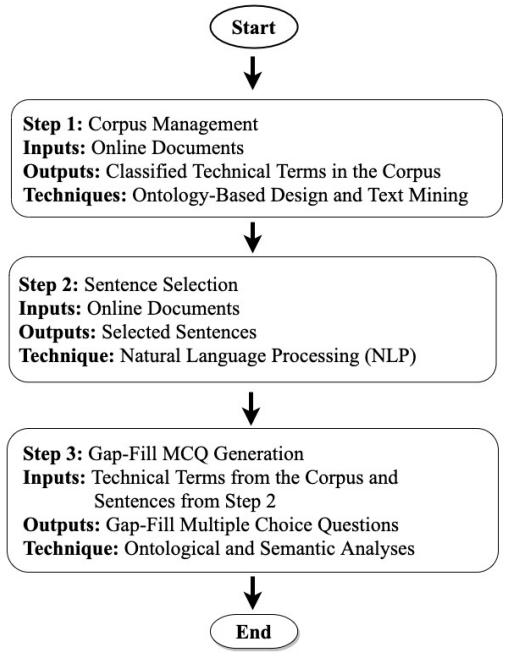

**Figure 2.** Three Main Steps of the Proposed Algorithm

***Step 1 – Corpus Management:*** A database specifically called *'corpus'* is constructed at this step to collect technical terms. These technical terms are actually those occurring in the documents describing technical concepts in a knowledge domain. Two techniques – ontology-based design and text mining – suggested in the previous research [38 - 39] are incorporated into the algorithm at this step to classify technical terms extracted from documents before collecting them in the corpus.

An ontology-based design is first used to specify concepts and concepts' relations in a knowledge domain. Figure 3 presents an ontology of Software Testing – the knowledge domain used to demonstrate an application of the algorithm. The focal concepts in this domain include Black-box Testing (BBT), White-box Testing (WBT), Boundary Value Analysis (BVA), Equivalence Partitioning (EP), Decision Table Testing (DT), Branch Coverage Testing (BR), and Statement Coverage Testing (STA). The relations between concepts (see Figure 3) are called *'generalization'* [42] meaning that BBT is a generic approach of the three specific software testing methods including BVA, EP, and DT. Likewise, WBT is a generic approach of the two specific testing methods including BR and STA.

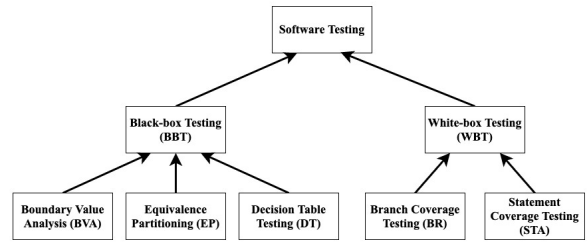

**Figure 3.** Concepts and Concepts' Relations in the Software Testing Domain

The ontology in Figure 3 is used as a guide to gather documents describing concepts in the domain. In this research, we used 7 concepts' titles – *'black-box testing'*, *'white-box testing'*, *'boundary value analysis'*, *'equivalence partitioning'*, *'decision table testing'*, *'branch coverage testing'*, and *'statement coverage testing'* in Google's search engine to gather online documents from web pages in the *'.edu'* domain.

**Table 1.** A Summary of Online Documents

| Concepts | #Documents | #Sentences | #Technical Terms | #Unique and Frequent Terms |
|---|---|---|---|---|
| BBT | 19 | 414 | 52 | 39 |
| WBT | 16 | 263 | 33 | 20 |
| BVA | 13 | 181 | 19 | 12 |
| EP | 16 | 347 | 26 | 12 |
| DT | 6 | 94 | 7 | 4 |
| BR | 22 | 108 | 13 | 10 |
| STA | 13 | 63 | 10 | 6 |

A summary of the gathered documents is in Table 1. Column *'#Documents'* in Table 1 presents the number of documents describing the concepts in Column *'Concepts'*. Column *'#Sentences'* presents the number of sentences in the documents, and Column *'#Technical Terms'* presents the number of technical terms occurring in the documents. In this research, we use the standard glossary published by the Software Testing Qualifications Board [43] to identify technical terms in the documents. Last, the number of key technical terms called *'unique terms'* and *'frequent terms'* is presented in Column *'#Unique and Frequent Terms'*.

This current research refers to the previous research studies [38 - 39] in defining three classes of technical terms – *'unique'*, *'frequent'*, and *'common'*. A unique term means the technical term occurring only in the documents of one specific concept. A frequent term means the technical terms mostly occurring in the documents describing a specific concept, and occasionally in the documents describing a relative concept. Based on the investigation of [38], both unique and frequent terms are basically employed to describe core characteristics of focal concepts in the documents. Different than unique and frequent terms, common terms are found generally in the documents of relative concepts in the same knowledge domain, and used to describe common ideas shared by relative concepts.

Especially for this current research, the algorithm is invented to find unique and frequent terms of seven concepts in the software testing domain. These terms would be collected in the corpus at Step 1, and later retrieved to use in Steps 2 – 3.

The Venn diagrams in Figures 4 [a] – [f] represent the number of unique, frequent and common terms occurring in the documents of paired concepts. Figure 4 [a] represents the number of unique and frequent terms of the two opposite testing approaches – BBT and WBT equal to 39 and 20, respectively. In the documents of BBT and WBT, we found 13 common terms shared by the two concepts.

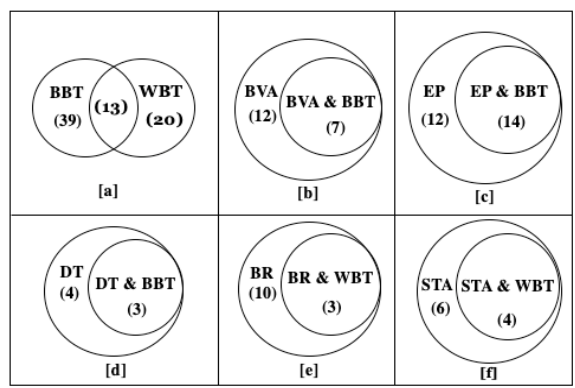

**Figure 4.** Occurrence of Unique and Frequent Terms in the Documents of Relative Concepts

The Venn diagrams in Figures 4 [b] – [d] represent the number of unique and frequent terms of the specific concepts derived from BBT including BVA, EP, and DT. In the documents of BVA, we found 12 unique and frequent terms specifically describing the characteristics of BVA, and 7 common terms describing the common ideas shared by BVA and BBT. Likewise, in the documents of EP and DT, we found 12 unique and frequent terms of EP, 4 unique and frequent terms of DT, 14 common terms shared by EP and BBT, and 3 common terms shared by DT and BBT.

The last two Venn diagrams (see Figures 4 [e] – [f]) represent the number of unique and frequent terms of the specific concepts derived from WBT including BR and STA. In the documents of BR and STA, we found 10 unique and frequent terms of BR, 6 unique and frequent terms of STA, 3 common terms shared by BR and WBT, and 4 common terms shared by STA and WBT, respectively.

***Step 2 – Sentence Selection:*** This step is for examining sentences suitable for being used as question phrases of gap-fill MCQs. A Natural Language Processing (NLP) technique is used to select sentences complying with the following 4 criteria:

- *Criterion 1: A sentence must contain a concept title, or tokens of a concept title.* A sentence that contains a concept title, e.g., *'black-box testing'*, *'white-box testing'*, *'statement coverage testing'*, *'branch coverage testing'*, or tokens of a concept title, e.g., *'black box'*, *'white box'*, *'statement coverage'*, *'branch coverage'* is suitable for using as a question phrase, because this type of a sentence is self-contained, and understandable by readers regarding the concept to which the question phrase refers.
- *Criterion 2: A sentence must contain a unique or a frequent term of a concept.* A unique or a frequent term is actually a key term representing the main idea of a concept. A sentence containing a unique or a frequent term is therefore worth being used to formulate a question.
- *Criterion 3: A sentence must be a main clause, not a subordinate clause.* A main idea is normally contained in the main clause, while a supplementary idea is in the subordinate clause [30]. The NLP technique used in this research would detect a subordinate clause from a conjunction at the beginning of the clause, e.g., *'in addition'*, *'however'*, *'otherwise'*, *'while'*, *'but'*. A clause beginning with a conjunction would be filtered out, and not used in formulating a question phrase.
- *Criterion 4: A sentence must be in the declarative type.* The NLP technique of this research is used to examine input sentences to ensure that the sentences are in form of *'subject + verb'* or *'subject + verb + object'*, so that there is no incomplete phrase or interrogative sentence used to formulate a question phrase.

***Step 3 – Gap-Fill MCQ Generation:*** Complete gap-fill MCQs are generated at this step. Figure 5 presents a complete structure of a gap-fill MCQ comprising two main parts – a question phrase and four answer choices.

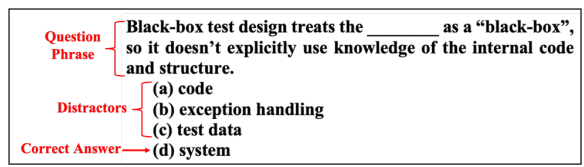

**Figure 5.** A Structure of a Gap-Fill Multiple Choice Question

A question phrase is constructed by removing a unique or a frequent term in the sentence, and replacing the spot of the term with a blank space. The term deprived from the sentence is then randomly put in a choice as a correct answer. The other three choices are distractors that actually are unique or frequent terms of other relative concepts of the concept mentioned in the question phrase. The unique and frequent terms in the distractor choices are retrieved from the corpus constructed at Step 1.

The working process of generating gap-fill MCQs is presented in Figure 6. The process starts from retrieving a sentence from the sentence collection built at Step 2. A unique or a frequent term in the sentence is pulled out, and replaced by a blank space. The term pulled out is used as a correct answer in one of the four choices (a) – (d). Finally, ontological and semantic analyses are applied to formulate three distractors.

An ontological analysis is applied to find relative concepts of the concept mentioned in the question phrase, and a semantic analysis is applied to find unique and frequent terms implying the same conceptual idea of the correct answer. Especially for the software testing domain, we classify the conceptual idea of the terms into four classes including thought, object, quality, and process. The working process at Step 3 would run continually until the sentence collection is empty.

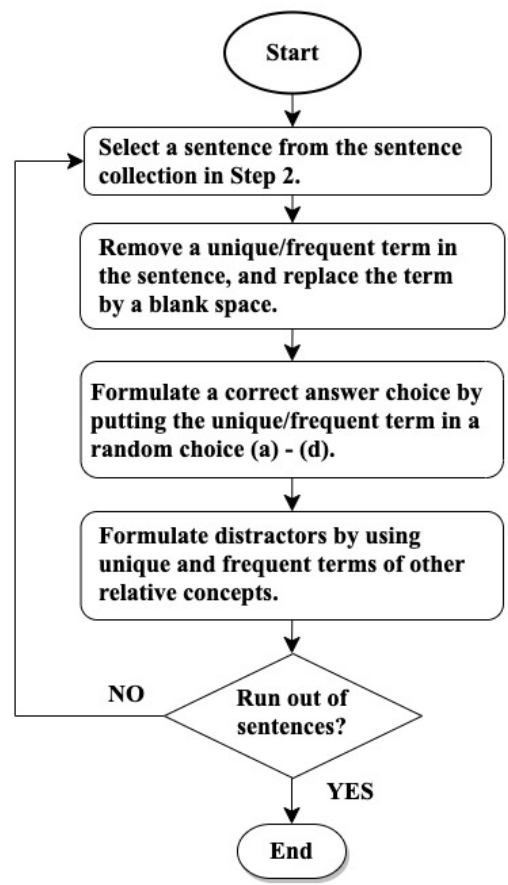

**Figure 6.** Process for Formulating Gap-Fill MCQs

[Figure 7 content - Gap-Fill MCQ examples]

**[a] Gap-Fill MCQ about Black-box Testing**

Black box testing is not concerned with the internal mechanisms of a _____; it focuses solely on the outputs generated in response to selected inputs and execution conditions.
(a) defect
(b) compiler
(c) decision table
(d) system

**[b] Gap-Fill MCQ about White-box Testing**

White box testing is mainly used for detecting logical errors in the _______.
(a) control flow graph
(b) code
(c) capture/replay tool
(d) compiler

**[c] Gap-Fill MCQ about Boundary Value Analysis**

Boundary Value Analysis is a selection technique in which test data are chosen to lie along boundaries of _______ (or output range) classes, data structures, procedure parameters, etc.
(a) domain
(b) functional requirement
(c) input domain
(d) equivalence partition

**[d] Gap-Fill MCQ about Statement Coverage Testing**

Even _______ that produce 100% statement coverage will MISS certain defects relating to control structures.
(a) test suites
(b) equivalence partitions
(c) decisions
(d) preconditions

**Figure 7.** Examples of Gap-Fill MCQs

The sample gap-fill MCQs in Figures 7 [a] – [d] are about the four software testing concepts including black-box testing, white-box testing, boundary value analysis, and statement coverage testing. The term *'system'* (see Figure 7 [a]) in choice (d) is the key term that makes the question phrase about black-box testing complete. The terms *'defect'*, *'compiler'*, and *'decision table'* are chosen from the corpus, and used as distractors in choices (a) – (c), because they are the unique and frequent terms of the relative concepts (i.e., white-box testing and decision table testing) of black-box testing, and also these three terms share the same conceptual idea about the *'object'* with the term in the correct answer choice (d).

The term *'code'* in the correct answer choice (b) (see Figure 7 [b]) is the term making the question phrase about white-box testing complete. The terms *'control flow graph'*, *'capture/replay tool'*, and *'compiler'* are chosen to be distractors in choices (a), and (c) – (d), because they conceptually mean objects used in the relative software testing concepts including branch coverage testing, black-box testing, and statement coverage testing, respectively.

The term *'input domain'* (see Figure 7 [c]) in the correct choice (c) makes the question phrase about boundary value analysis fulfilled. The terms in the correct answer, and distractors conceptually refer to thoughts and expectations in different software testing concepts. The terms in the distractors are unique and frequent terms of the concepts about black-box testing and equivalence partitioning.

The term *'test suite'* in choice (a) (see Figure 7 [d]) is the key term of the question phrase about statement coverage testing. This key term and the terms in the distractor choices share the conceptual idea about logical thoughts for preparing data in software testing. The three terms in the distractors are the unique and frequent terms of the concepts about equivalence partitioning, white-box testing, and black-box testing.

At the last step, the algorithm would take a role of converting the terms in the four multiple choices to be in the grammatical form required at the position of the filling gap in the question phrase. As presented in Figure 7 [a], for example, all choices (a) - (d) would be checked to ensure they are nouns not beginning with a vowel as the position before the filling gap is an article *'a'*. Likewise, all multiple choices in Figure 7 [d] would be checked to ensure they are plural nouns compatible with the plural verb *'produce'* in the question phrase.

# 4. Simulation Study and Results

**Table 2.** Outputs from Software Built Based on the Proposed Algorithm

| Testing Concepts | #Generated Question Phrases | #Valid Question Phrases | #Gap-Fill MCQs |
|---|---|---|---|
| BBT | 29 | 25 | 3,020 |
| WBT | 29 | 28 | 6,088 |
| BVA | 8 | 8 | 2,174 |
| EP | 6 | 6 | 1,424 |
| DT | 1 | 1 | 99 |
| BR | 9 | 9 | 2,469 |
| STA | 3 | 3 | 864 |
| **Total** | **85** | **80** | **16,138** |

We conducted a simulation of this research by building software to implement the proposed algorithm through Steps 1 - 3. The outputs from running software at Step 1 are the technical terms classified as unique, frequent, and common terms as presented in Figure 4.

Table 2 presents outputs from software running at Steps 2 – 3. Column *'Testing Concepts'* presents seven software testing concepts stated in the question phrases. Column *'#Generated Question Phrases'* presents the number of question phrases generated by the software at Step 2. These question phrases were validated by 3 validators based on a 2 out of 3 agreement, that is, a question phrase would be considered valid if 2 out of 3, or all 3 validators agreed that the question phrase was understandable, and also informative enough for the readers to predict a term missing from the question phrase. The number of valid question phrases is presented in column *'#Valid Question Phrases'*. Out of 85 question phrases, 80 (94.12%) were valid. This high percentage does show quality of the algorithm in accurately selecting sentences suitable for formulating question phrases.

Column *'#Gap-Fill MCQs'* presents the number of complete gap-fill MCQs formulated by combining question phrases with multiple choices. In validating gap-fill MCQs automatically generated by software, we asked the 3 validators to specifically examine effectiveness of distractors. If 2 out of 3, or all 3 validators felt they needed to justify between at least two possible answer choices, the entire set of distractors in the gap-fill MCQ would be considered valid.

We found the algorithm very successful in producing distractor sets. The validators agreed that there was at least one distractor in every gap-fill MCQ that made the validators hesitate to make a choice. Moreover, we found that the proposed algorithm had strength in producing a large number of varied gap-fill MCQs. As presented in columns *'#Valid Question Phrases'* and *'#Gap-Fill MCQs'* in Table 2, around 80 question phrases could be used to produce more than 16 thousand gap-fill MCQs. This happened because one question phrase could be incorporated with numerous sets of multiple choices formulated from varied unique and frequent terms in the corpus.

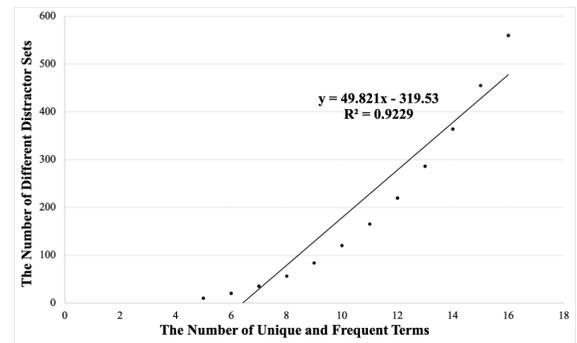

**Figure 8.** Correlation between the Number of Unique and Frequent Terms and the Number of Different Distractor Sets

Given a key term used as a correct answer in a gap-fill MCQ, the software could retrieve unique and frequent terms of relative concepts to build varied sets of distractors. For example, the key term *'system'* in a gap-fill MCQ about black-box testing was mapped with other 7 unique and frequent terms of the concepts about white-box testing, boundary value analysis, equivalence partitioning, and decision table testing, and these 7 terms were randomly selected by the software to produce different 35 sets of distractors.

Based on data in this simulation study, we found a key term in a gap-fill MCQ could be mapped with 5 to 16 unique and frequent terms suitable for building distractors. Figure 8 shows a high correlation, $r(26) = 0.96$, p-value $< .00001$, between the number of unique and frequent terms and the number of different distractor sets built from the unique and frequent terms.

We can conclude from this high correlation that unique and frequent terms are actually the key for formulating varied sets of distractors in gap-fill MCQs. The more unique and frequent terms that are used for describing concepts in a knowledge domain, the more varied gap-fill MCQs can be produced by the proposed algorithm.

Based on the results of the simulation, we can also indicate that the technical terms essentially convey the ideas asked in the gap-fill MCQs. Figure 9 is the diagram presenting the 4 groups of gap-fill MCQs representing the conceptual ideas about *'THOUGHT'*, *'OBJECT'*, *'QUALITY'*, and *'PROCESS'*. Details of the diagram follow.

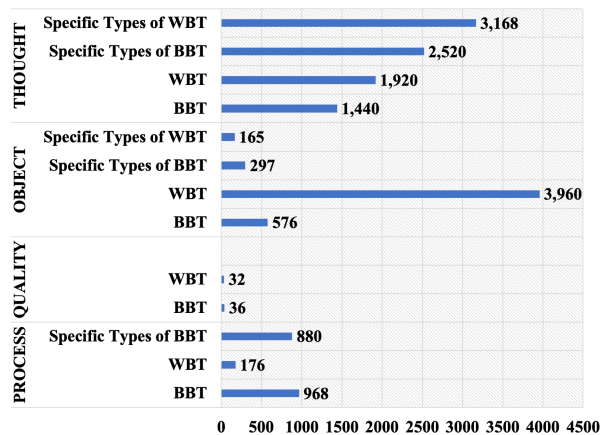

**Figure 9.** The Four Conceptual Idea Groups of Gap-Fill MCQs

- The largest number of gap-fill MCQs was found in group *'THOUGHT'*. The gap-fill MCQs in this group essentially asked about the specific ideas of testing software by using BBT and WBT approaches, and the specific methods of BBT and WBT including BVA, EP, DT, BR and STA.

- We found group *'OBJECT'* mostly contained gap-fill MCQs asking about the objects, e.g., tools, documents, and output/input data used in the WBT approach. Some other questions asked about the objects used in BBT and BBT's specific testing methods.

- Gap-fill MCQs in group *'QUALITY'* is rather small, because there are not too many technical terms conveying the ideas about quality of software. The gap-fill MCQs in this group are about quality of software tested by BBT and WBT approaches.

- Last, the gap-fill MCQs found in group *'PROCESS'* are about the testing processes of WBT, BBT and BBT's specific testing methods including BVA, EP and DT.

## 5. Discussion and Future Work

As demonstrated by the simulation, the proposed algorithm has strength in producing a tremendous number of gap-fill MCQs from about a hundred documents. We therefore suggest to apply this algorithm to build a question pool for use in a knowledge expert system.

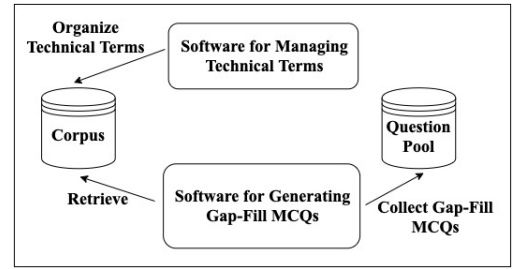

**Figure 10.** Building a Question Pool at Design Time

Figure 10 presents how the algorithm can be applied at design time to automatically produce gap-fill MCQs collected in the question pool. The module *'Software for Managing Technical Terms'* is code implemented based on Step 1 of the proposed algorithm used for classifying technical terms in the corpus. The module *'Software for Generating Gap-Fill MCQs'* is code implemented based on Steps 2-3 of the proposed algorithm. This software would retrieve technical terms from the corpus to produce gap-fill MCQs being collected in the question pool.

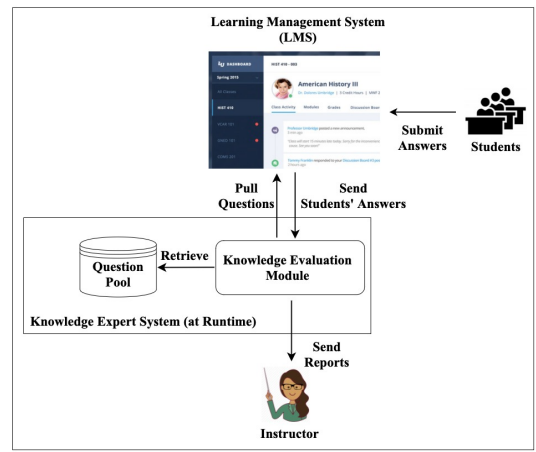

**Figure 11.** Knowledge Expert System at Runtime

At run time, the question pool would work with the knowledge evaluation module as a part of a knowledge expert system (see Figure 11). The knowledge expert system, from another view, is a plug-in module taking a role of evaluating students' learning performance for a Learning Management System (LMS). When the knowledge expert system is enabled, the LMS would pass gap-fill MCQs from the knowledge expert system to students. When the students submit their answers, the LMS would pass the students' answers to the knowledge evaluation module in the knowledge expert system. After evaluating students' answers, the knowledge evaluation module would send an instructor a report about students' learning performance.

## 6. Conclusion

In this paper, we propose an algorithm for automatically generating gap-fill multiple choice questions (MCQs). This algorithm focuses on constructing questions that can evaluate knowledge in academia, especially in scientific domains. Based on the proposed algorithm, key technical terms are mainly used to formulate question phrases, and multiple choices comprising one correct answer, and three distractors. In the simulation study, we found the algorithm could effectively produce a tremendous number of gap-fill MCQs covering all core concepts in the knowledge domain. Finally, we imply from the research findings that this algorithm is promising if applied as an automated question generator in a knowledge expert system used for evaluating knowledge in online learning.

## 7. Acknowledgement

This research was jointly supported by National Science and Technology Development Agency (NSTDA) of Thailand [grant code FDA-CO-2562-8930-TH]; the Educational Exchange Program under the Collaboration between National Research Council of Thailand (NRCT) and Indian Council of Social Science Research (ICSSR); and Rajamangala University of Technology Phra Nakhon, Thailand.